\documentclass[a4paper,conference]{IEEEtran}

\usepackage{xcolor}
\usepackage{graphicx}
\usepackage{amsmath}
\usepackage{amssymb}
\usepackage{acronym}
\usepackage{array}
\usepackage{placeins}
\usepackage{siunitx}
\usepackage{epstopdf}
\usepackage{textcomp}

\usepackage[caption=false,font+=footnotesize]{subfig}
\renewcommand{\figurename}{Fig.}
\newsavebox{\measurebox}

\makeatletter
\def\ps@IEEEtitlepagestyle{
  \def\@oddfoot{\mycopyrightnotice}
  \def\@evenfoot{}
}
\def\mycopyrightnotice{
  {\footnotesize
  \begin{minipage}{\textwidth}
  \centering
  \copyright~2021 IEEE.  Personal use of this material is permitted.  Permission from IEEE must be obtained for all other uses, in any current or future media, including reprinting/republishing this material for advertising or promotional purposes, creating new collective works, for resale or redistribution to servers or lists, or reuse of any copyrighted component of this work in other work.
  \end{minipage}
  }
}

\begin{document}
\newacro{vae}[VAE]{variational autoencoder}
\newacro{adc}[ADC]{analog-to-digital converter}
\newacro{fmcw}[FMCW]{frequency modulated continuous wave}
\newacro{rai}[RAI]{range-angle image}
\newacro{rdi}[RDI]{range-doppler image}
\newacro{da}[DA]{domain adaptation}
\newacro{hvac}[HVAC]{heating, ventilation, and air conditioning}
\newacro{mm-wave}[mm-wave]{millimeter-wave}
\newacro{cfar}[CFAR]{constant false alarm rate}
\newacro{rcs}[RCS]{radar cross section}
\newacro{dbscan}[DBSCAN]{density-based spatial clustering of applications with noise}
\newacro{lstm}[LSTM]{long short-term memory}
\newacro{rf}[RF]{radio frequency}
\newacro{pll}[PLL]{phase-locked loop}
\newacro{vco}[VCO]{voltage-controlled oscillator}
\newacro{mti}[MTI]{moving target indication}
\newacro{mvdr}[MVDR]{minimum variance distortionless response}
\newacro{esprit}[ESPRIT]{estimation of signal parameters via rotational invariance techniques}
\newacro{music}[MUSIC]{multiple signal classifier}
\newacro{oscfar}[OS-CFAR]{ordered statistic constant false alarm rate}
\newacro{cfel}[CFEL]{complex frequency extraction layer}
\newacro{kl}[KL]{Kullback-Leibler}
\newacro{dcnn}[DCNN]{deep convolutional neural network}

%
\title{Radar Image Reconstruction from Raw ADC Data using Parametric Variational Autoencoder with Domain Adaptation}

%

\author{\IEEEauthorblockN{{Michael~Stephan\IEEEauthorrefmark{1}\IEEEauthorrefmark{2}\IEEEauthorrefmark{3}, Thomas~Stadelmayer\IEEEauthorrefmark{1}\IEEEauthorrefmark{2}\IEEEauthorrefmark{3}, Avik~Santra\IEEEauthorrefmark{2}, Georg~Fischer\IEEEauthorrefmark{1}, Robert~Weigel\IEEEauthorrefmark{1}, Fabian~Lurz\IEEEauthorrefmark{1}}}
\IEEEauthorblockA{\textit{\IEEEauthorrefmark{2}Infineon Technologies AG}, Neubiberg, Germany \\
\IEEEauthorrefmark{1} Friedrich-Alexander-University Erlangen-Nuremberg, Erlangen, Germany\\
Email: \{thomas.stadelmayer, avik.santra\}@infineon.com}, \{michael.stephan, georg.fischer, robert.weigel, fabian.lurz\}@fau.de\\
\IEEEauthorrefmark{3}equal contribution}



\maketitle

\begin{abstract}
This paper presents a parametric \acl{vae}-based human target detection and localization framework working directly with the raw \acl{adc} data from the \acl{fmcw} radar. 
We propose a parametrically constrained \acl{vae}, with residual and skip connections, capable of generating the clustered and localized target detections on the \acl{rai}. Furthermore, to circumvent the problem of training the proposed neural network on all possible scenarios using real radar data, we propose \acl{da} strategies whereby we first train the neural network using ray tracing based model data and then adapt the network to work on real sensor data. This strategy ensures better generalization and scalability of the proposed neural network even though it is trained with limited radar data. We demonstrate the superior detection and localization performance of our proposed solution compared to the conventional signal processing pipeline and earlier state-of-art deep U-Net architecture with \acl{rdi}s as inputs.

\end{abstract}
\acresetall


\IEEEpeerreviewmaketitle

\section{Introduction}
Human detection and localization is essential for smart home applications that leverage this information to automatically control lighting and \ac{hvac} systems to significantly reduce energy consumption in residential, commercial or mall settings. Several studies have shown that energy consumption can be significantly reduced by 25\%-75\% \cite{garg2000smart} by sensing the number of people and their respective locations. Furthermore, human detection and localization is essential for monitoring and maintaining social distancing guidelines, which is of high importance especially under the current corona crisis. There are several sensors that can enable reliable human detection and localization in indoor environments, including vision-based sensors. However, unlike vision-based sensors, radar-based human detection and localization offers a solution that is robust to lighting conditions and preserves privacy \cite{peng2016portable, wang2014hybrid, santra2018short}. Furthermore, \ac{mm-wave} radars have small form factors, resulting from integrated silicon and antenna-in-package, that can be easily mounted and aesthetically integrated on the operating device. Radar-based human detection facilitates people counting and density estimation \cite{radarBook, choi2017people}, and is also prerequisite for human activity classification \cite{lin2018human, vaishnav2020continuous} that can enable further intelligent control of appliances. Moreover, radar-based detection finds use in automotive in-cabin occupancy sensing \cite{alizadeh2019low}.           
Traditional, the \ac{fmcw} radar signal processing pipeline for human detection and localization involves translating the raw \ac{adc} data along fast-time, i.e., time index within a chirp, and slow-time, i.e., time index across chirps, to range-Doppler domain by 1D FFTs along both axes one after the other. The fast-time axis transforms to range and slow-time to Doppler axis. Reflections from a single human target cause multiple detections on the range and Doppler dimension due to high bandwidth chirps used in \ac{mm-wave} radars and micro-Doppler components arising from human body parts under motion respectively. Once the \ac{rdi} is generated for all the virtual channels, they are fed into a beamforming algorithm to transform the channel data to explicit angle information. Depending on the algorithm and processing pipeline, a \ac{rai} or range-angle-Doppler 3D data cube is generated. The \ac{rai} is then fed into a \ac{cfar} detector to ensure target detection under the Neymann-Pearson criteria. This is followed by clustering to ensure that multiple detections from a single target are grouped together as a single target. Additionally, spurious reflections are rejected as outliers.
Human target detection and localization in indoor environments using conventional signal processing is particularly challenging since they lead to ghost targets on the \ac{rai} due to multi-path reflections on the human body from walls, chairs, and furniture. The complex relation between range, Doppler, target \ac{rcs} and interactions among targets can lead to missed detections on one hand, such that strong reflections from a close-by human can almost occlude the reflection from a farther human target, and spurious targets on the other hand, as reflections from a single human can manifest as several split targets \cite{stephan2019radar}. Further, several hyper-parameters in signal processing algorithms, such as the scaling factor, the guard - and training-window size in \ac{cfar} or the minimum number of clustering points and distance separation in \ac{dbscan}, result in inaccurate target detections in the form of ghost targets or missed targets \cite{stephan2020deep}.
Recently, several papers \cite{fuchs2020automotive, wang2019study, zhang2020object, major2019vehicle, 8835792} have been proposed that aim to replace the traditional detection and clustering algorithms using deep autoencoder architectures and have demonstrated much better detection performances compared to traditional signal processing approaches. In \cite{stephan2019radar}, a deep U-Net with residual connections has been proposed to perform detection and clustering on the \ac{rdi}, whereas in \cite{stephan2020deep} a complex deep U-Net with residual connections has been proposed to transform \acp{rdi} from different channels to a \ac{rai} and perform detection and clustering on that domain. In \cite{fuchs2020automotive}, a deep autoencoder has been proposed for interference mitigation and amplitude-phase reconstruction in \acp{rai} for outdoor radars. In \cite{major2019vehicle}, a modified deep U-Net with \ac{lstm} cells has been presented for automotive target detections in polar range-angle domain. In \cite{zhang2020object}, an autoencoder architecture has been suggested for transforming \acp{rdi} from multiple channels to \acp{rai} for detecting automotive targets.
However, these works use \acp{rdi} across channels as input for their proposed neural network, which may not be effective in fully exploiting the potentials of deep neural networks that are, in principle, capable of extracting feature images implicitly. Moreover, 1D FFTs along fast-time followed by 1D FFTs along slow-time to generate \acp{rdi} are the computationally most intensive step of the conventional radar signal processing. Furthermore, such hybrid signal processing and deep learning designs are not efficient for embedded solutions as the FFTs need to be performed by the microcontroller or general processor before passing the \acp{rdi} into a deep learning accelerator. To address these issues, in this paper we propose a parametrically constrained autoencoder, where the first layer is constrained to parametric functions that enable filtering operation along fast-time - slow-time simultaneously and resembles the (range and Doppler) frequency separability property of FFTs. SincNet-based parametric 1D CNNs have first been proposed for audio signal processing \cite{ravanelli2018speaker}. Similar parametric 2D \acp{dcnn} for radar-based human activity classification have been proposed in \cite{stadelmayer2020short}. For scalability and generalization of the proposed deep neural network, a large training dataset, where data is collected under several different conditions and configurations, is required. Nevertheless, training the network with a dataset comprising such varied scenarios and configuration is practically implausible. To overcome such challenges, we, in this paper, propose a \ac{da} strategy to first train the proposed architecture using synthetic data generated through ray tracing models under numerous configurations followed by network adaptation to real sensor data, which is generally limited. 

\section{Radar System Design}

\subsection{Radar Chipset}
\begin{figure}%
	\centering
	\subfloat[Photo of the radar chipset]{{\includegraphics[width=0.4\linewidth]{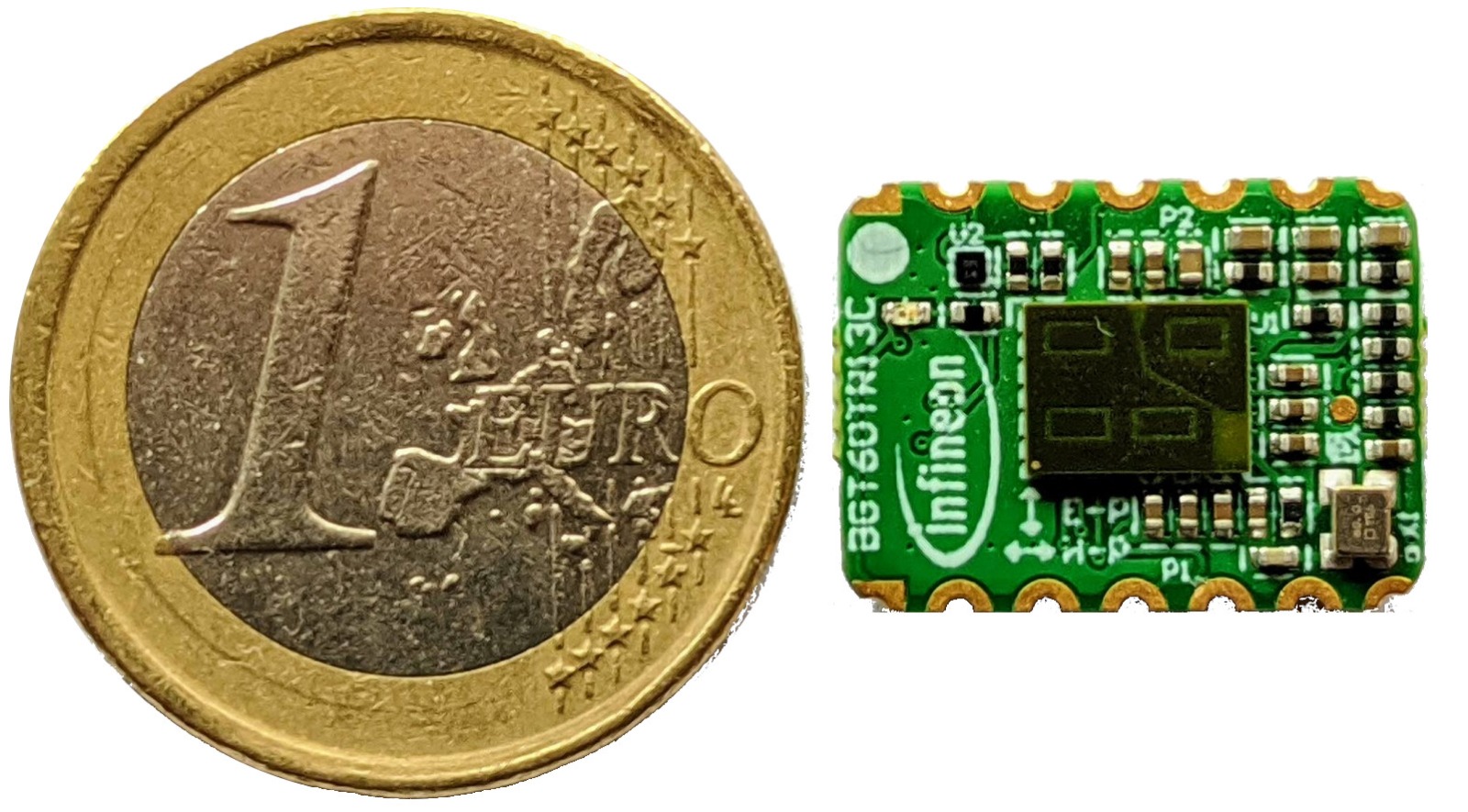} }} 
	\subfloat[Typical \ac{fmcw} block diagram]{{\includegraphics[width=0.6\linewidth]{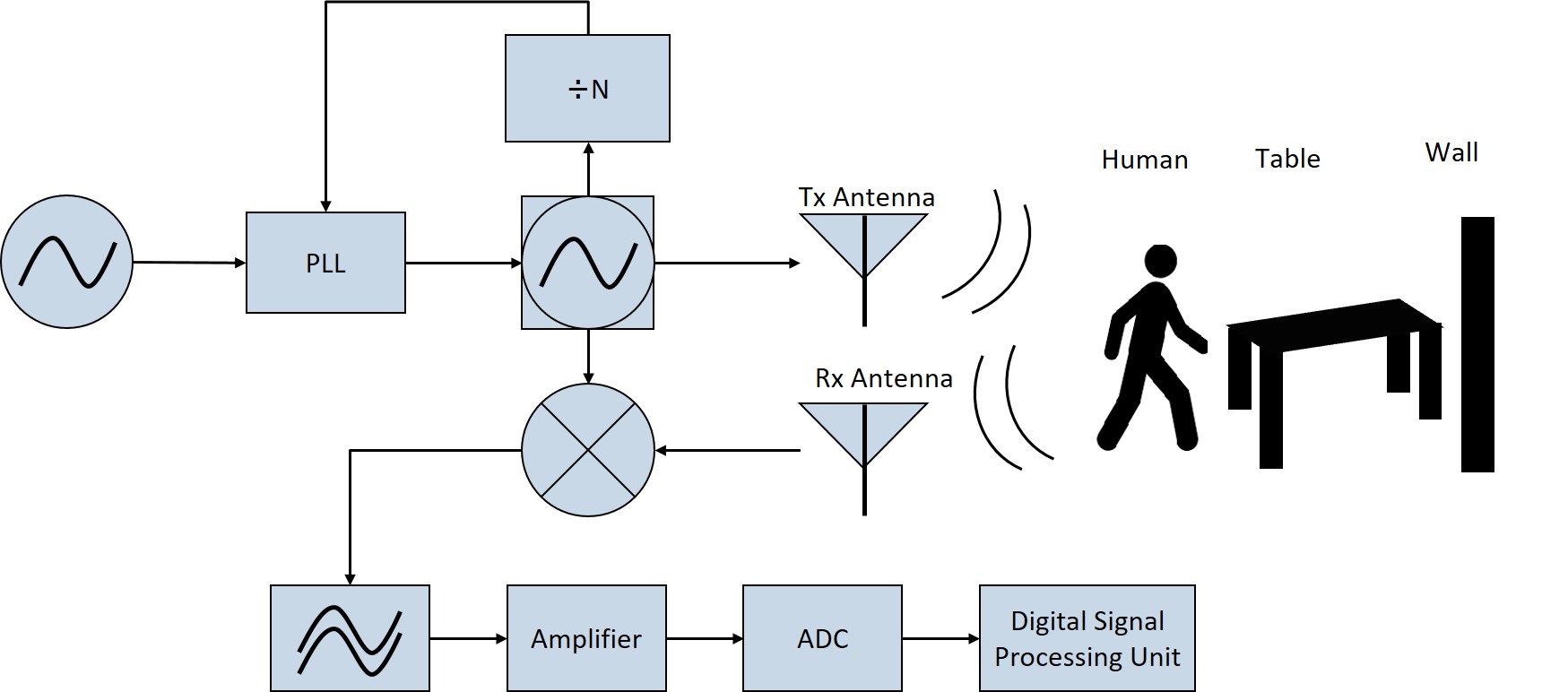} }} 
	%
	\caption{(a) \emph{Infineon}'s \emph{BGT60TR13C} 60-GHz radar sensor. (b) Functional block diagram of \ac{fmcw} radar \ac{rf} signal chain depicting 1TX, 1RX channel}%
	\label{fig:radar_chipset}%
\end{figure}

In the paper \emph{Infineon}'s \emph{BGT60TR13C} \ac{fmcw} radar chipset is used. An image of the chipset and the analog \ac{rf} signal chain for one transmit and one receiving antenna is depicted in Fig.~\ref{fig:radar_chipset}. In the transmit part, the \ac{pll}, with a reference frequency of \SI{80}{\mega\hertz}, regulates the \ac{vco}. The radar is able to generate highly linear chirps covering a frequency range between \SI{57}{\giga\hertz} and \SI{64}{\giga\hertz} within a configurable chirp duration by adjusting the divider value and an additional tuning voltage ranging from \SI{1}{\volt} to \SI{4.5}{\volt}. The backscattered signal is received by the receiving antenna, then mixed with a replica of the transmitted signal and afterwards low pass filtered in order to obtain the intermediate or beat frequency signal. Further, the signal is sampled with a sampling frequency of \SI{2}{\mega\hertz} by the \SI{12}{bit} \ac{adc}. Since the radar has three receiving antennas there are three such receive paths. The antennas are ordered in an L-shape, and therefore two antennas each can be used for azimuth and elevation angle estimation. In this paper, only the two receiving antennas that are necessary to estimate the azimuth angle are used. The relative offset of the used antennas is $d = \SI{2.5}{\milli\meter}$, and therefore half the wavelength of the transmitted signal. Moreover, the sensor covers a field of view of \SI{70}{\degree} in elevation and \SI{120}{\degree} in azimuth angle dimension.\\

\subsection{Signal Model and System Configuration}
\label{subsec:signal}
The mixed and low pass filtered signal of a single chirp assuming a backscatterer at distance $d_{\text{n}}$ is defined as
\begin{align}
    S_{\text{n}, \text{ft}}(d_{\text{n}}, t_\text{ft}) &= A\cos(2\pi(\frac{2Bd_{\text{n}}}{cT}t_\text{ft} + \frac{2f_0d_{\text{n}}}{c} + \frac{B}{2T}(\frac{2d_{\text{n}}}{c})^2)) \notag\\
    &\approx A\cos(2\pi(\frac{2Bd_{\text{n}}}{cT}t_{\text{ft}} + \frac{2f_0d_{\text{n}}}{c}))
\end{align}
where $A$ is the voltage amplitude, $B$ is the chirp bandwidth, $d_\text{n}$ is the radial distance of target $n$ to the radar, $f_0$ is the center frequency, c is the speed of light, $T$ is the chirp time and $t_\text{ft}$ is the fast-time.
Due to low velocities in indoor environments, the scene within a chirp is quasi-static. Moreover, the distances are very small for indoor applications, and therefore the quadratic term can be neglected. However, when observing the signal over multiple chirps, i.e. slow-time, the displacement of a moving target can be detected. The received, mixed and low pass filtered signal over slow-time is defined as
\begin{align}
   & S_{\text{n}}(d_{\text{n}}, t_{\text{ft}}, t_{\text{st}}) = \notag \\
   & A\cos(2\pi(\frac{2B(d_{\text{n}} + v_{\text{n}}t_{\text{st}})}{cT}t_{\text{ft}} + \frac{2f_0(d_{\text{n}} + v_{\text{n}}t_{\text{st}})}{c}))
\label{equ:sig_2d}
\end{align}
where the parameters represent the same physical sizes as before and additionally $t_\text{st}$ is the slow-time axis and $v_\text{n}$ is the velocity of target $n$. Relative to the absolute distance $d_{\text{n}}$ the movement $\delta d = v_{\text{n}}t_{\text{st}}$ is negligible small. However, since the phase is periodical to the wavelength $\lambda = c/(2f_0)$, which is about \SI{5}{\milli\meter} for a center frequency of $f_0 = \SI{60}{\giga\hertz}$, small displacements still induce a noticeable phase shift. As a result \eqref{equ:sig_2d} can be approximated as
\begin{align}
    &S_{\text{n}}(d_{\text{n}}, t_{\text{ft}}, t_{\text{st}}) \approx \notag \\ &A\cos(2\pi(\frac{2Bd_{\text{n}}}{cT}t_{\text{ft}} + \frac{2f_0(d_{\text{n}} + v_{\text{n}}t_{\text{st}})}{c})).
\end{align}

Moreover the signal received by an antenna is the superposition of the backscattered signal of multiple targets. The final signal representation of a single antenna used in this paper is therefore defined as
\begin{equation}
    S = \sum_{\text{n} \in \text{N}} S_{\text{n}}(d_{\text{n}}, t_{\text{ft}}, t_{\text{st}})
\end{equation}
where N is a set of targets.

The signal over multiple receiving antennas differs by a phase shift defined as
\begin{equation}
    \phi = \frac{2\pi d \sin(\theta)}{\lambda}
\end{equation}
where d is the distance between the receiving antennas, $\theta$ is the relative azimuth angle of the target and $\lambda$ is the wavelength. By analyzing the phase offset, the angle of arrival can be estimated.
The radar chipset in the paper was configured to send out chirps starting from $f_{\text{min}} = \SI{58}{\giga\hertz}$ up to a maximum frequency of $f_{\text{max}} = \SI{62}{\giga\hertz}$ within a chirp time of $T_{\text{c}} = \SI{261}{\micro\second}$. The received signal is sampled 256 times with a sampling frequency of $f_{\text{s}} = \SI{2}{\mega\hertz}$. The total chirp bandwidth $B = \SI{4}{\giga\hertz}$ results in a range resolution of $\delta r = \SI{3.75}{\centi\meter}$ and therefore a maximum range $R_{\text{max}}$ of \SI{4.8}{\meter} can be observed. Moreover a data frame consists of 32 chirps with a chirp repetition time $T_{\text{PRT}}$ of \SI{520}{\micro\second}. As a result a Doppler resolution of $\delta v_{\text{max}} = \SI{1.25}{\meter\per\second}$ is obtained.

\subsection{Classical Processing Pipeline}

For indoor target detection and localization, multiple processing steps have to be performed. First, a 2D FFT is applied on the raw \ac{adc} data. Then, a \ac{mti} filter is applied to mitigate static targets. Afterwards, an imaging angle estimation method is applied and finally, a target detection followed by a clustering method has to be performed.

\subsubsection{2D FFT}
As discussed, the range and Doppler information of one or more targets in the field of view is encoded in the frequency composition of the signal along the fast- and slow-time dimension, respectively. Hence, the signal is traditionally preprocessed using a two staged FFT. First, a 1D FFT is applied along the mean removed chirps. Afterwards, a second 1D FFT is applied on the FFT processed signal across a set of chirps. In both cases, a window function is applied. This processing results in a complex-valued \ac{rdi}. To mitigate static targets and antenna leakage, \ac{mti} filtering is applied over multiple \acp{rdi}.

\subsubsection{Angle Estimation}
After the signal was disassembled into different range-Doppler bins, an angle estimation can be performed. As stated in sec. \ref{subsec:signal} the angle of arrival is defined by the phase shift over multiple receiving antennas. However, due to noise and other disturbances evaluating only the phase difference leads to inaccurate results. There exist advanced angle estimation approaches such as the \ac{mvdr} beamformer, the \ac{esprit}, and the \ac{music} method. We use the \ac{mvdr} method to obtain the \ac{rai}.

\subsubsection{Object Detection}
From the \ac{rai}, one or multiple targets have to be extracted. This can either be done by simple thresholding or by applying more advanced techniques like \ac{oscfar}. Afterwards, the \ac{rai} is given in a binary manner, where zero stands for no target, and one stands for a detected target in a range angle bin. When a person is walking, the signal is not only reflected by the body, but also by the moving limbs which can generate a small isolated signal close to the main target. In order to detect this as a single target, we apply \ac{dbscan} for clustering. Based on the resulting clusters, the location of one or more human targets can be extracted.

\begin{figure}%
	\centering
	\subfloat[Traditional signal processing]{{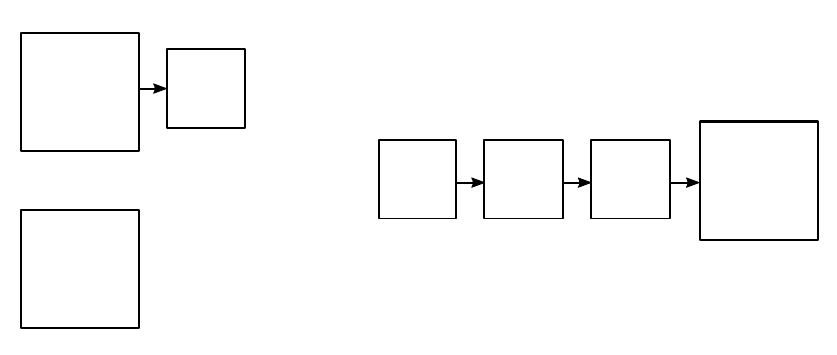 }} \\
	\subfloat[Proposed parametric VAE]{{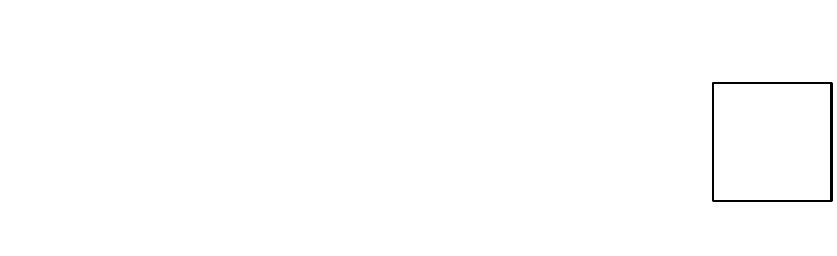 }} 
	%
	\caption{(a) Traditional signal processing pipeline (b) proposed solution using a parametric VAE}%
	\label{fig:processing_pipelines}%
\end{figure}

\section{Proposed Solution}
In this paper, we propose a deep learning based human target detection and localization solution that replaces the entire traditional signal processing chain as described in the previous chapter. The raw \ac{adc} data is fed into the neural network, enabling a more efficient workflow due to the fact that the data does not have to be preprocessed on a digital signal processor. Instead, the raw data can directly be transferred from the sensor to specialized accelerator hardware. Until now, almost all state of art deep learning based radar systems use preprocessed input data. It turns out that the system often gets stuck in a local minimum when training it on raw \ac{adc} data. However, we propose using a parametric layer at the beginning of the neural network for range and Doppler feature extraction. This is then followed by a \ac{vae} architecture. Moreover, we propose training the neural network on synthetic radar data and then use \ac{da} in order to adapt the neural network to real world data.

\subsection{Complex Frequency Extraction Layer}
Processing the data using a 2D FFT directly unveils range and Doppler information but in turn is compute intense and the number of sampling points limits its accuracy. Instead, we propose using a \ac{cfel} for range and Doppler feature extraction. This is a parametric layer, which means that the filter kernels are given by a function defined by a finite set of parameters. In the proposed \ac{cfel} the filter kernels are defined as
\begin{align}
    f_{\text{M, N}}(f_{\text{ft}}, f_{\text{st}}; m, n) = e^{j2\pi f_{\text{ft}}m/f_{\text{s}}^{\text{ft}}} e^{j2\pi f_{\text{st}}n/f_{\text{s}}^{\text{st}}}
\end{align}
where m and n are the sample indices, M and N the filter lengths and $f_{\text{s}}^{\text{ft}}$ and $f_{\text{s}}^{\text{st}}$ are the sampling frequencies in fast- and slow-time respectively. Moreover, $f_{\text{ft}}$ and $f_{\text{st}}$ are the learnable hyperparameters that define the filter kernels. These hyperparameters also define the frequencies that the filter kernel extracts from the signal. Additionally, to create a set of filter kernels the number of filters in fast- $N_\text{ft}$, as well as in slow-time $N_\text{st}$, has to be given. Although each filter is applied and trained independently, the output channels are reshaped to a two dimensional matrix of size $N_\text{ft} \times N_\text{st}$ to obtain a similar representation as a \ac{rdi}.
Learning the filter frequencies enables the possibility, unlike in an FFT, to analyze the signal composition in some frequency areas in more detail than in others. In order to enable equal training, the hyperparameters are normalized. When the filters are created using the set of harmonic frequencies, the output of the \ac{cfel} equals a 2D FFT.
\begin{figure}%
	\centering
	\subfloat[Real part]{{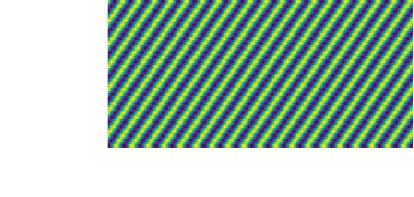 }}
	\subfloat[Imaginary part]{{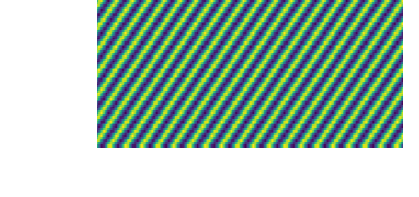 }}\\
	\subfloat[Frequency response]{{
\begingroup%
  \makeatletter%
  \providecommand\color[2][]{%
    \errmessage{(Inkscape) Color is used for the text in Inkscape, but the package 'color.sty' is not loaded}%
    \renewcommand\color[2][]{}%
  }%
  \providecommand\transparent[1]{%
    \errmessage{(Inkscape) Transparency is used (non-zero) for the text in Inkscape, but the package 'transparent.sty' is not loaded}%
    \renewcommand\transparent[1]{}%
  }%
  \providecommand\rotatebox[2]{#2}%
  \newcommand*\fsize{\dimexpr\f@size pt\relax}%
  \newcommand*\lineheight[1]{\fontsize{\fsize}{#1\fsize}\selectfont}%
  \ifx\svgwidth\undefined%
    \setlength{\unitlength}{120.47244094bp}%
    \ifx\svgscale\undefined%
      \relax%
    \else%
      \setlength{\unitlength}{\unitlength * \real{\svgscale}}%
    \fi%
  \else%
    \setlength{\unitlength}{\svgwidth}%
  \fi%
  \global\let\svgwidth\undefined%
  \global\let\svgscale\undefined%
  \makeatother%
  \begin{picture}(1,0.47058824)%
    \lineheight{1}%
    \setlength\tabcolsep{0pt}%
    \put(0.23392156,0.05676467){\color[rgb]{0,0,0}\makebox(0,0)[lt]{\lineheight{1.25}\smash{\begin{tabular}[t]{l}\scriptsize{0}\end{tabular}}}}%
    \put(0.87018551,0.05547859){\color[rgb]{0,0,0}\makebox(0,0)[lt]{\lineheight{1.25}\smash{\begin{tabular}[t]{l}\scriptsize{255}\end{tabular}}}}%
    \put(0.46678137,0.0141496){\color[rgb]{0,0,0}\makebox(0,0)[lt]{\lineheight{1.25}\smash{\begin{tabular}[t]{l}\scriptsize{range bins}\end{tabular}}}}%
    \put(0.16903689,0.12919122){\color[rgb]{0,0,0}\makebox(0,0)[lt]{\lineheight{1.25}\smash{\begin{tabular}[t]{l}\scriptsize{0}\end{tabular}}}}%
    \put(0.13941841,0.41636702){\color[rgb]{0,0,0}\makebox(0,0)[lt]{\lineheight{1.25}\smash{\begin{tabular}[t]{l}\scriptsize{31}\end{tabular}}}}%
    \put(0.1175427,0.13570071){\color[rgb]{0,0,0}\rotatebox{89.89370481}{\makebox(0,0)[lt]{\lineheight{1.25}\smash{\begin{tabular}[t]{l}\scriptsize{Doppler bins}\end{tabular}}}}}%
    \put(0,0){\includegraphics[width=\unitlength,page=1]{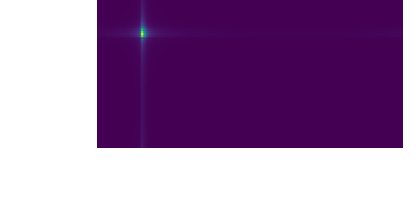}}%
  \end{picture}%
\endgroup%
 }} 
	\caption{(a) Real part, (b) imaginary part and (c) frequency response of an exemplary \ac{cfel} filter kernel}%
	\label{fig:filter_kernel}%
\end{figure}
The real and imaginary part, as well as the frequency response of an exemplary filter kernel, is shown in Fig.~\ref{fig:filter_kernel}. It can be seen that the frequency response is a single sharp peak. Thus, applying a set of these filter kernels can be seen as sampling the underlying signal in frequency domain at variable positions.

\subsection{Variational Autoencoder}
\label{subsec:VAE}
A general idea of encoder-decoder-based structures is to compress some input data into a smaller latent space so that the desired output data can still be obtained from this compressed representation. Ideally, the latent space should capture some high-level features, like the presence/absence of real targets and ghost targets, which are then used in the decoder to highlight targets and remove ghost targets. In practice, the data representation chosen by the encoder network may not be as interpretable, in case of strong overfitting, single input examples may even be mapped to some specific numbers without any high-level meaning. 
A way to achieve continuity in the latent space and reduce overfitting is to use a \ac{vae} instead.
In a \ac{vae}, the input is not encoded into a single point in latent space, but into a distribution over the latent space. This is achieved, by sampling from normal distributions with mean vectors $\mu$, and standard deviations $\sigma$, the values of which are learned by the network. The decoder then learns the mapping from these distributions to the desired outputs. During inference, only the values of $\mu$ are used, each input example now corresponds to a deterministic latent vector, where similar inputs should correspond to similar latent vectors.
The vectors $\mu$ and $\sigma$ are the outputs of two fully connected layers in the network. As the sampling operation itself is not differentiable, a reparametrization trick is used to move the sampling operation out of the backpropagation-path. Partly in order to avoid the network to set the learned variances to zero, and therefore sample from dirac distributions, the \ac{kl} divergence from the learned distributions defined by $\mu$ and $\sigma$ and Gaussian distributions is added as a regularization term to the overall loss.  

\subsection{Domain Adaptation}
\label{subsec:DA}

As is often the case for radar data, the amount of labeled data to train our neural network on is rather limited. While augmentation techniques exist for radar data too, compared to image processing, they are much less powerful, and a lot of possible input scenarios will, therefore, not be covered by the training data. To still steer our network towards general detection capabilities, we use domain adaptation techniques inspired by soft-parameter sharing multi-task learning, as it was shown in \cite{duong-etal-2015-low}, for dependency parsing in linguistics.

We first train a network with the same architecture on labeled synthetic point target data, covering the whole output range-angle space. We then initialize the network for the real target data with these pretrained weights and add a weight difference regularization term to our loss function, as shown in \eqref{eq:weight_reg}.
\begin{equation}
    \label{eq:weight_reg}
    L_{DA} = \sum_{i \in T} \frac{||w_i - w_{i0}||^2 + ||b_i - b_{i0}||^2}{||w_{i0}||^2 + ||b_{i0}||^2}
\end{equation}
Using \eqref{eq:weight_reg}, we slightly punish any divergence of the weights to the weights of the network trained on the synthetic dataset. Here, $w_i$ and $b_i$ are the weights and biases for the $i$th convolutional layer, $T$ is the set of convolutional layers in our models, while $w_{i0}$ and $b_{i0}$ are the weights and biases for the same convolutional layers trained on the synthetic data. The idea behind this approach is to have the neural network learn a general angle of arrival estimation function on the synthetic data, which should also generally be needed to detect real targets in the range angle space in addition to the ghost target removal and other functions. In effect, this also reduces the experienced overfitting due to the weight regularization.


\section{Architecture \& Learning}

	\subsection{Architecture}
\begin{figure*}
	\includegraphics[width=\textwidth]{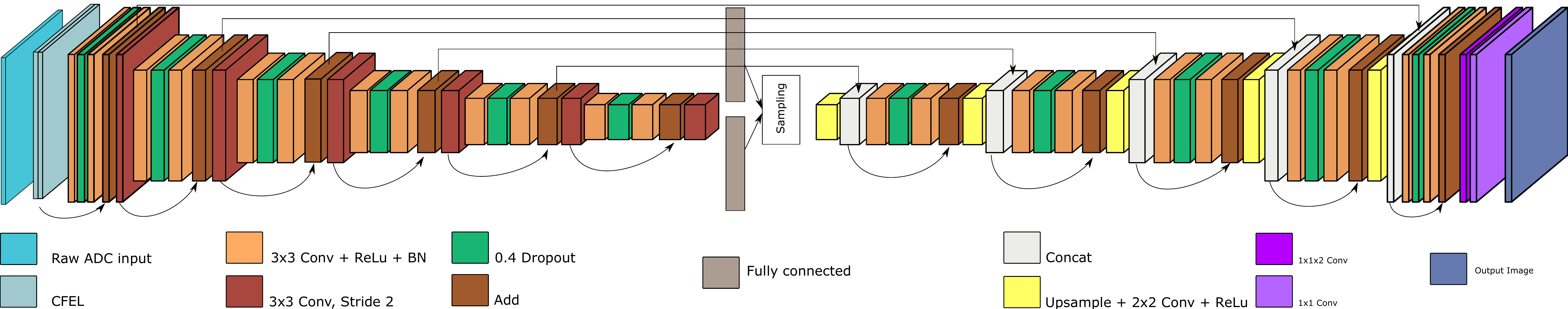}
  	\caption{Proposed parametric \ac{vae} architecture for human target detection from raw \ac{adc} data}
  	\label{fig:full_network}
\end{figure*}
Fig.\ref{fig:full_network} shows the full proposed architecture for human target detection from raw \ac{adc} data. The neural network input is the two-channel raw \ac{adc} data from two receiving antennas of the radar chip. The first layer is the \ac{cfel}. 
The \ac{cfel} is initialized so that the center frequencies form a uniform grid over the range-Doppler space. The kernel size in fast-time direction is chosen to be 256 and 32 in slow-time dimension. The initial learnable frequencies in fast-time direction are defined by 128 equally spaced frequencies from 0 to $f_{s}^{ft}/2$, whereas the initial learnable frequencies in slow-time are given by 32 equally spaced frequencies ranging from 0 to $f_{s}^{st}$. In doing so, 128 different positive range bins and 32 Doppler bins, including positive and negative velocities, are created. The 2D kernels are then generated by combining each fast-time frequency with each slow-time frequency resulting in a total set of 4096 kernels. This setup equals a 2D FFT using 128x32 samples. Each kernel is then trained on its own, so its underlying frequencies are independent of other kernels, which allows a maximal flexible and efficient 2D frequency extraction. The same \ac{cfel} layer is applied independently to the raw \ac{adc} inputs from the two antennas. The output is then of size $128\times32\times2\times2$, where the third dimension describes the two antennas, and the last dimension the channel dimension, where the two channels are the real and the imaginary output part of the \ac{cfel}. The encoder part of the network consists of six blocks with the same structure. Each block consists of a $3\times3\times1$ convolution, followed by a 0.4 dropout and another $3\times3\times1$ convolution. The input to the block is then added to the convolution output, after undergoing a $1\times1\times1$ convolution to match the channel dimension. Lastly, the feature map size is reduced by a factor of two in both direction through a $3\times3\times1$ convolution with a stride of $2\times2\times1$. Between each encoder block, the number of channels is increased by a factor of 1.6, the number is rounded down to the nearest integer. Following the encoder are two fully connected layers, representing the mean-vector and the variance-vector, and a sampling layer to draw from the distribution defined by the fully connected layers. Each fully connected layer has 140 neurons, corresponding to the flattened input, except for the antenna dimension. 
The decoder part also consists of five blocks of the same structure and of similar structure to the encoder blocks. The input to each such block is the output from the previous block. First it is upsampled by a factor of two, followed by a $2\times2\times1$ convolution. The convolution output is then concatenated with the output of the add layer of the corresponding encoder block. This is again followed up by two $3\times3\times1$ convolutions with a 0.4 dropout layer in between. Up until the last add layer in the decoder block, the same network is basically applied separately to the two antenna channels. The information from the two antennas is only brought together in the $1\times1\times2$ convolution just after the last add layer. The last 1x1 convolution is done to reduce the output channel size to one.

	\subsection{Loss Function}
	The final loss function, as shown in \eqref{eq:total_loss}, is a combination of the focal loss, the \ac{da} regularization term, and the \ac{kl} divergence loss term.
	\begin{equation}
	    \label{eq:total_loss}
	    loss = L_{FL} + \beta \cdot L_{DA} + \theta \cdot L_{KL}
	\end{equation}
	The focal loss, as described in \cite{FocalLoss}, is used due to the heavy class imbalance of targets and non-targets in our labeled data. Due to the varying number of targets and the varying target sizes, a constant class weighting is not sufficient. Using the focal loss, a higher emphasis is put on miss-classified output pixels during training. The equation for the focal loss is shown in \eqref{eq:focal_loss}.
	\begin{equation}
	\label{eq:focal_loss}
	\begin{aligned}
    	    &L_{FL} = \alpha_t \cdot (1-p_t)^\gamma \, \cdot \, \text{log}(p_t)\\
            &p_t = \left\{\begin{aligned}
                                  p \hspace{1cm} &\text{if} \;\; y = 1\\
                                  1-p \hspace{0.5cm} &\text{otherwise}
                                  \end{aligned}\right. \hspace{0.5cm}
    	\alpha_t = \left\{\begin{aligned}
                                  &1  &\text{if} \;\; y = 1\\
                                  &\alpha  &\text{if} \;\; y = 0
                                  \end{aligned}\right.
    \end{aligned}
	\end{equation}
In \eqref{eq:focal_loss}, $p_t$ describes the probability that a pixel was correctly classified. The parameter $\alpha$ controls the initial class weighting. If the value for $\gamma$ is chosen to one, the loss is equal to the crossentropy. For higher values of $\gamma$, more weight is put on miss-classified examples. We chose $\gamma = 2$ , and $\alpha = 0.25$ for our experiments. The other two components of the loss function as shown in \eqref{eq:total_loss} are the \ac{kl} divergence term and the domain adaptation term as described in sections \ref{subsec:VAE} and \ref{subsec:DA}. Both terms are multiplied with some factors to scale their total loss contribution, $\beta = 0.0001$ for the \ac{da}-loss, and $\theta = 0.1$ for the \ac{kl}-loss.
When training the network on the synthetic data, $\beta$ is set to 0.

\section{Results \& Discussion}

	\subsection{Dataset}
\label{subsec:Dataset}
For training our network with domain adaptation as presented earlier, we need two different datasets. One is the synthetic dataset for learning the general angle of arrival estimation, and the other is our real target dataset. To make it easier for the network to work with both these datasets, we normalize our synthetic and our real input data. Specifically, we map the lowest \ac{adc} input value to zero and the highest to one for each input image. 
	\subsubsection{Domain Adaptation}
	Using the phased array toolbox from Matlab, a radar with the same antenna positions and configurations as described in \ref{subsec:signal} was simulated. The space is discretized in 128 range positions from \SI{3.75}{\centi\meter} up to \SI{4.8}{\meter} and 32 azimuth angle positions ranging from -\SI{50}{\degree} to \SI{50}{\degree}. On each point in the discrete space, a point target was placed, and the radar signal was simulated using ray tracing. The received signal is then dechirped and downsampled in order to obtain the IF signal before it is stored on the hard drive.
	\subsubsection{Target Domain}
\begin{figure}[!htb]
	\label{fig:setup}
	\includegraphics[width=\columnwidth]{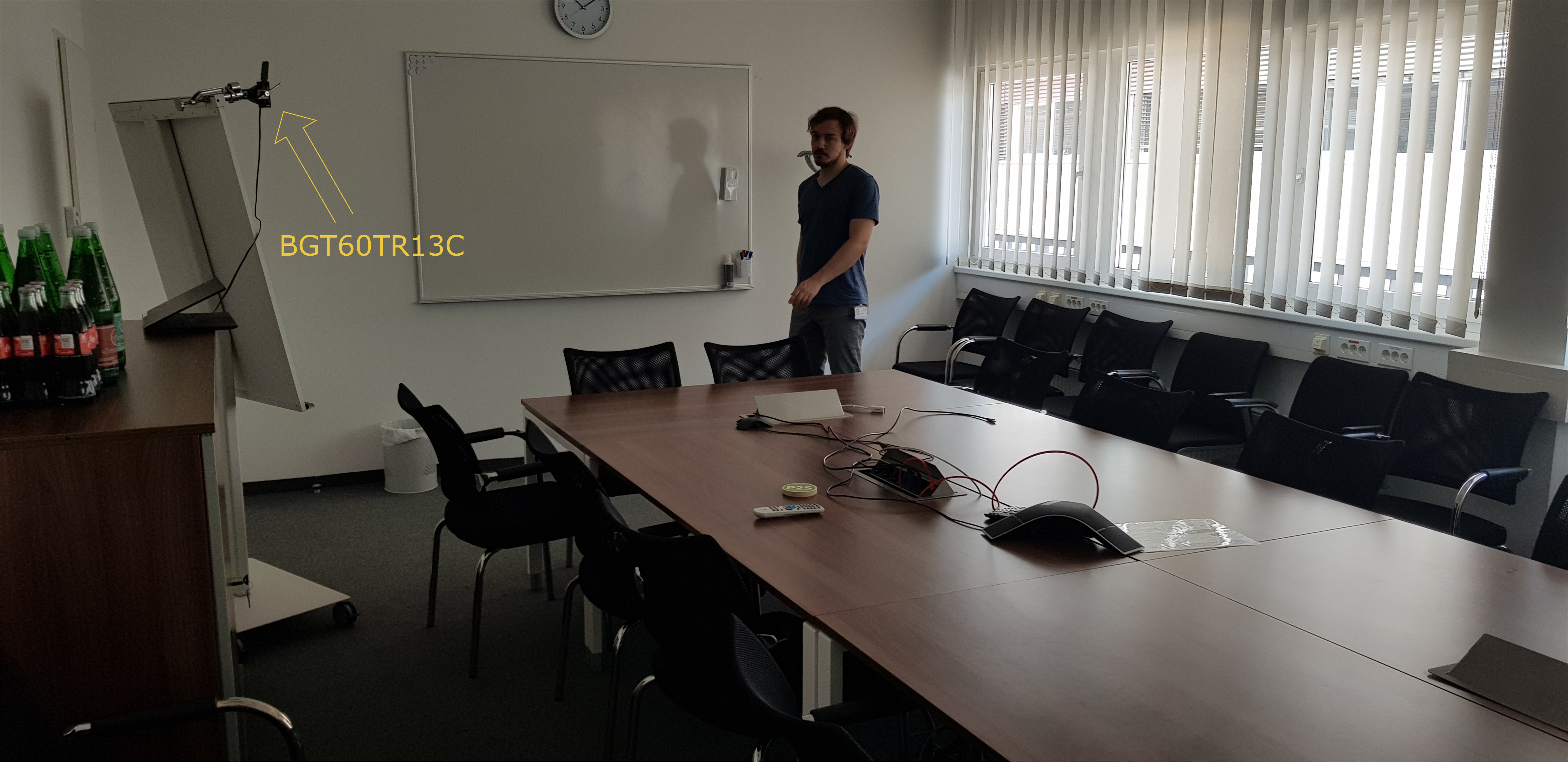}
  	\caption{Recording environment}
\end{figure}

To gather our real target dataset, we recorded a person walking around in a typical conference room with a camera and a radar sensor. The labeled data was then created using the traditional signal processing chain and by crosschecking its outputs with the camera data and manually removing any ghost targets or adding missed detections to the labels. The described recording scene is shown in Fig.\ref{fig:setup}.
Like this, 1200 labeled examples were created, 1000 for the training/validation set, and 200 for the test set.
These one target measurements are then taken as a base to create multi-target measurements, up to four targets, by superposition of different one target measurements. To reduce overfitting, it is necessary to employ some augmentation on the one target measurements. We slightly shift the measurements and labels in terms of range via multiplication of complex exponentials $\text{exp}^{j 2 \pi f t}$, where $f$ describes the applied frequency shift, across the fast-time dimension of our \ac{adc} data. In total, we then have 15000 training/validation examples and 1000 examples for testing.
	\subsection{Reconstruction Results}
   \begin{figure}[!ht]
     \subfloat[Time domain data\label{subfig-1:raw}]{%
       \includegraphics[width=0.3\linewidth]{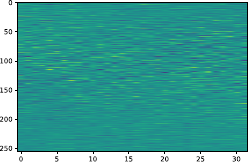}
     }
     \subfloat[RDI\label{subfig-1:rdi}]{%
       \includegraphics[width=0.3\linewidth]{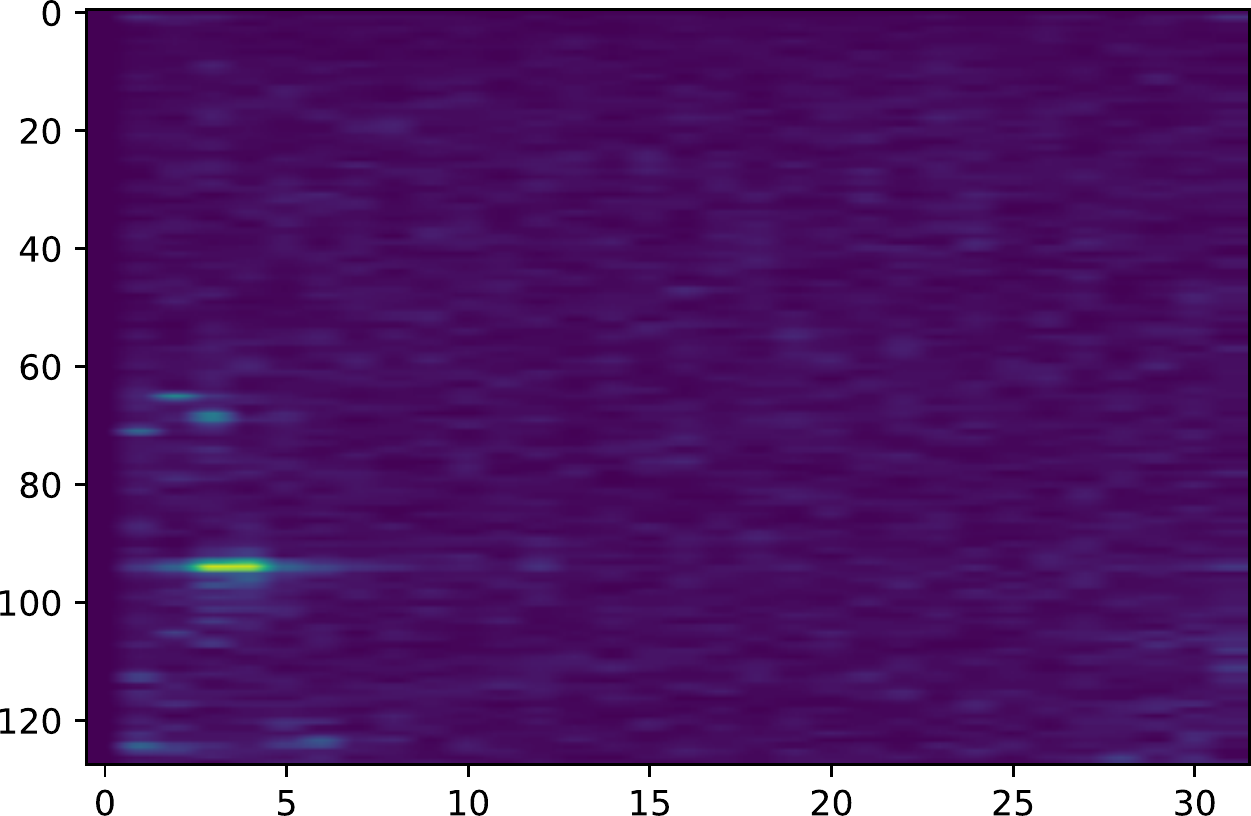}
     }
    \subfloat[Features after \ac{cfel}\label{subfig-1:wave_out}]{%
       \includegraphics[width=0.3\linewidth]{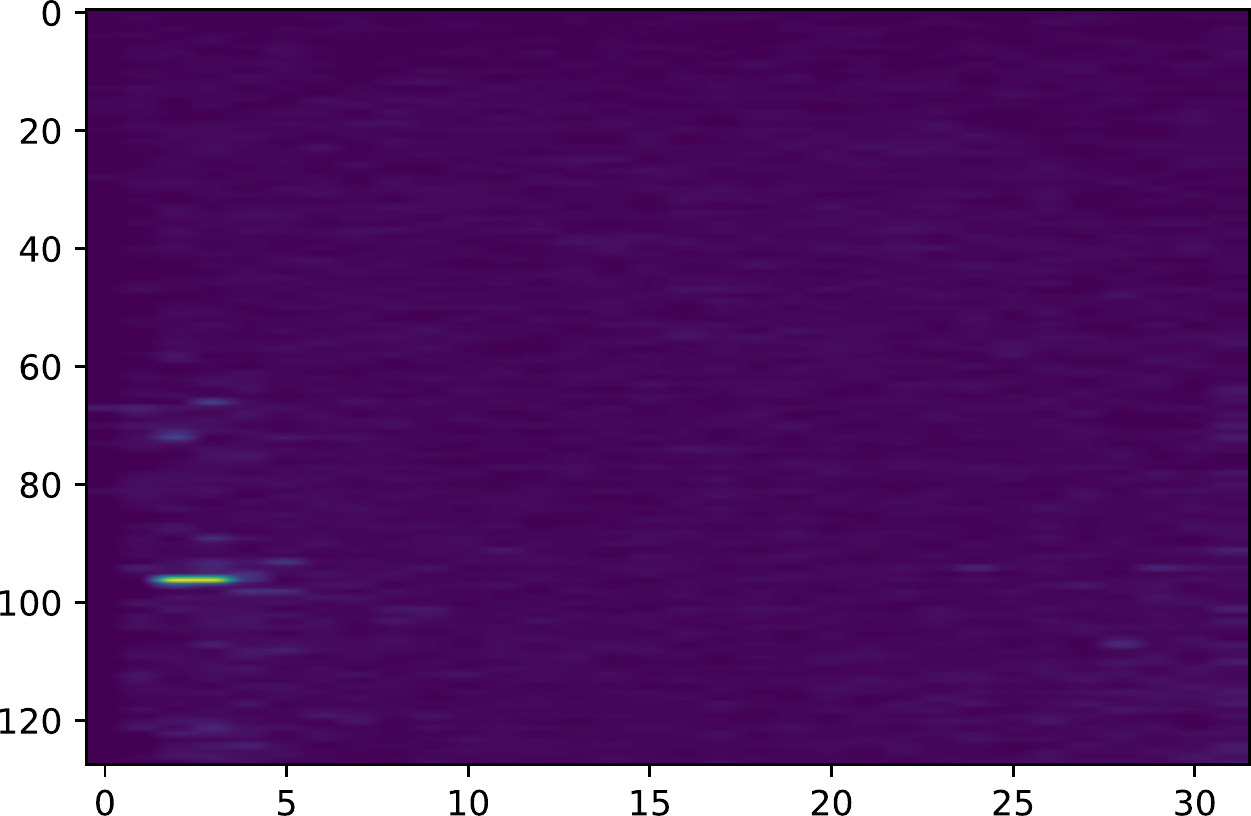}
     }
	\caption{Time domain data and corresponding output of the \ac{cfel} layer}
	\label{fig:reconstruction}
     \end{figure}
     
It is proposed to feed raw time-domain data into the neural network and extract meaningful features using a parametric \ac{cfel}. As a result, the \ac{cfel} replaces the preprocessing, which typically involves \ac{mti} filtering and a 2D FFT. As a comparison, the time domain signal, the absolute values of the preprocessed \ac{rdi}, and the absolute values of the output of the already trained \ac{cfel} of the same scene are shown in Fig.~\ref{fig:reconstruction}. The x-axis goes along slow-time (a) or Doppler (b) (c) and the y-axis along fast-time (a) or range (b) (c) direction. It can be seen that it is hardly possible to obtain information from the time domain signal itself, but after preprocessing, the target gets unveiled as a peak in the range-Doppler domain. After initialization, the features after the \ac{cfel} layer look the same as the \ac{rdi}, but during training the frequencies that are analyzed or extracted are optimized. Thus, if the application requires, it is possible to get a higher frequency resolution in more meaningful frequency areas and less resolution in less meaningful areas. This is an advantage, especially in classification tasks. However, if all frequency regions are of equal interest for the application, the analyzed frequencies are distributed over the whole domain. In Fig.~\ref{fig:reconstruction} it can be seen that in the \ac{cfel}, the target information is successfully extracted from the time domain signal. Both the preprocessed \ac{rdi} as well as the output of the \ac{cfel} show consistent results.

	\subsection{Detection Results}
We evaluate our proposed approach on a set of 800 test measurements, consisting of one to four moving targets, with 200 examples per target number. The 200 one-target measurements are real measurements that are not within the training/validation set. The 2-4 target measurements were created from these one-target measurements as described in section \ref{subsec:Dataset}. We evaluate our model in terms of F1-score by looking at the number of missed detections and false alarms in each output image. We do this by comparing the center of masses of the clusters in the neural network output to the corresponding labels. We do a minimum bipartite matching in terms of euclidean distance between the clusters and count a target as detected only if the cluster in the network output is within a certain range of the cluster in the labeled image, specifically $37.5$ cm.

\begin{figure*}[!ht]
     	\includegraphics[width = \linewidth]{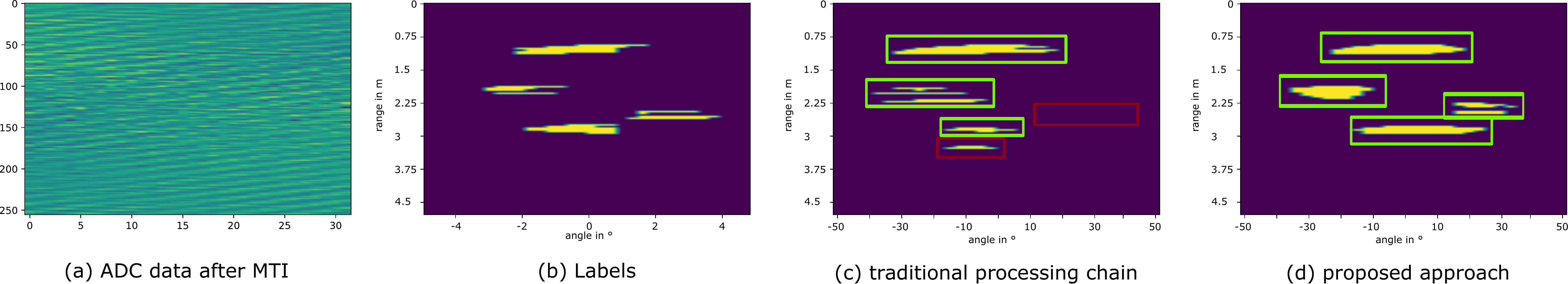}
     	\caption{(a) \ac{adc} data after \ac{mti}, (b) Labels indicating the true positions, (c) Processed \ac{rai} using the traditional chain , (d)  Processed \ac{rai} using proposed approach wherein all targets are detected accurately.}
     	\label{fig: split}
     \end{figure*}

\begin{figure*}[!ht]
     	\includegraphics[width = \linewidth]{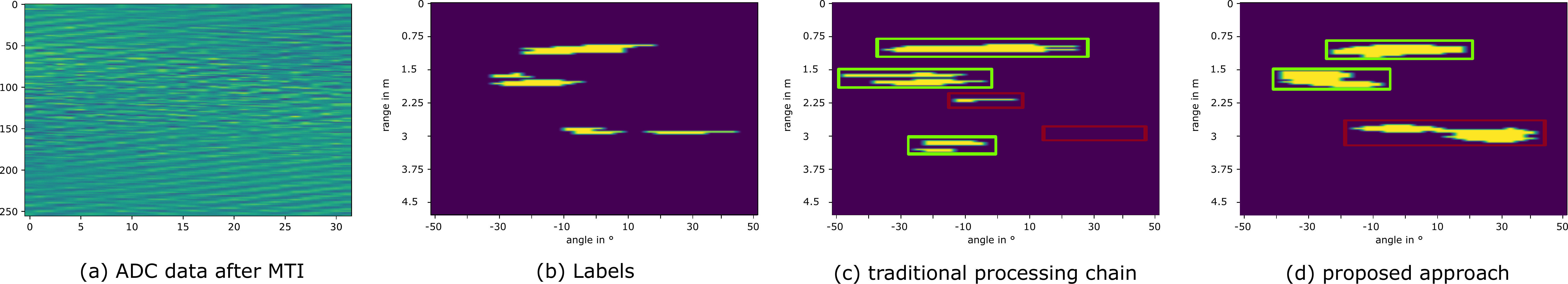}
     	\caption{(a) \ac{adc} data after \ac{mti}, (b) Labels indicating the true positions, (c) Processed \ac{rai} using the traditional chain , (d)  Processed \ac{rai} using proposed approach wherein all targets are detected accurately.}
     	\label{fig:NNmerge}
     \end{figure*}

	\begin{table}[!ht]
		\centering
		\caption{Comparison of the detection performance of the traditional pipeline with the AE architecture from \cite{stephan2020deep}, and the proposed method}
		\setlength{\tabcolsep}{1pt}
		\begin{tabular}{| l | l | l | l |}
			\hline
			Approach & Description & F1-Score & Model Size\\ \hline
			Traditional & \ac{oscfar} with \ac{dbscan}  & $ 0.61$ & - \rule{0pt}{2ex}\\  
			AE & AE with complex \ac{rdi} as input \cite{stephan2020deep}& $0.77$ & $1.23$ MB \rule{0pt}{2ex}\\
            \textbf{VAE} & \ac{vae} with \ac{cfel} layer and \ac{da} & $\boldsymbol{0.80}$ & $1.71$ MB \rule{0pt}{2ex}\\
			\hline		
		\end{tabular}
		\vspace{0.1cm}
		\label{table:approaches}
	\end{table}
	Table \ref{table:approaches} shows the results of the traditional signal processing approach, the method from our old paper, and for our new proposal in terms of F1-score on the described test-set. Both deep learning based methods show clear improvement over the traditional chain for the \ac{rai} detection task. The new method shows the best results on the test-set, mostly due to changes made in the network architecture. Some examples illustrating the model performance compared to the traditional processing chain are shown in Fig. \ref{fig: split}, \ref{fig:NNmerge}. Here, the (b) are the labels, (c) shows the output of the traditional processing chain, and (d) the output of the proposed approach. All correct detections are marked with a green box, while missed detections or false alarms have a red box instead. In Fig. \ref{fig: split}, the proposed approach shows all targets detected at the correct positions, while the traditional processing chain has one missed detection and one false alarm due to a target split. In Fig. \ref{fig:NNmerge}, there is a ghost target and a missed detection in (c), while two targets were detected as one with the neural network due to just using a simple clustering to avoid any additional processing overhead.

	\subsection{Discussion}    
\begin{figure}
        \def\svgwidth{\linewidth}
        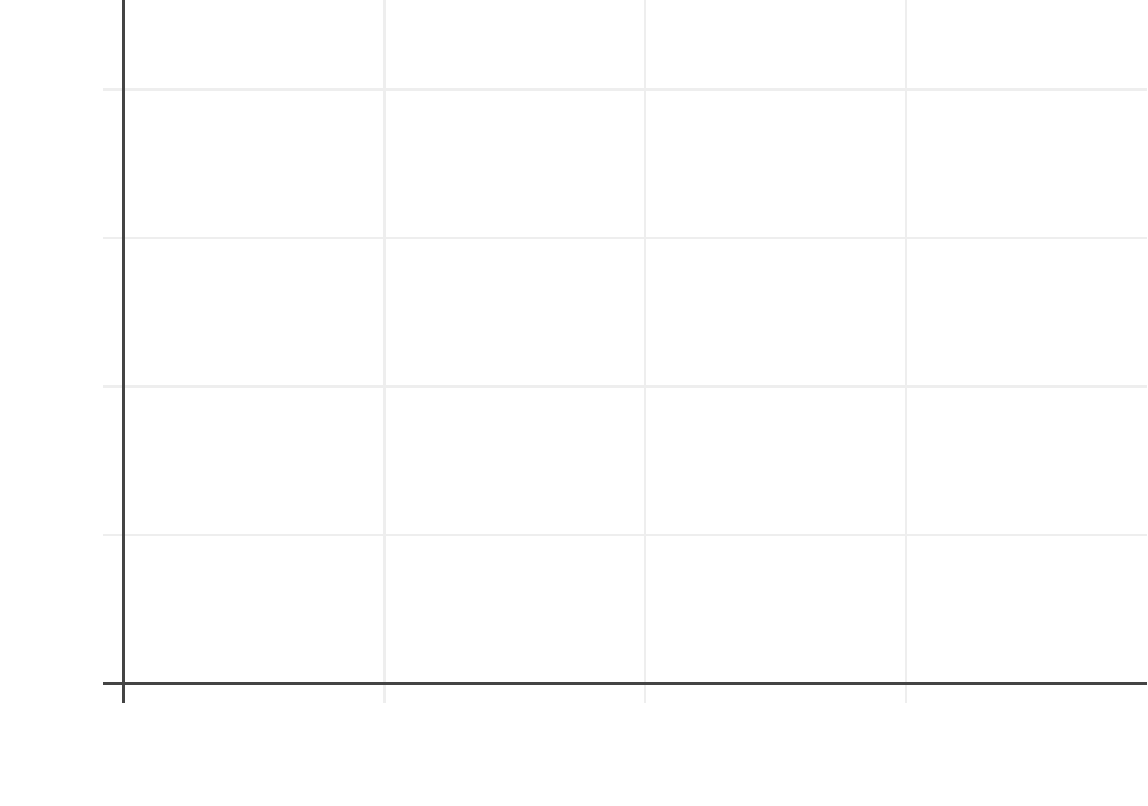
        \caption{Weight divergence from the synthetic model}
        \label{fig:weight_div}
    \end{figure}

		The proposed method works directly with the \ac{adc} data as input, making any additional preprocessing blocks unnecessary, therefore reducing the memory and compute requirements from the preprocessing, which is not factored in the model size in table \ref{table:approaches}. Furthermore, it is more robust to new input data due to the \ac{vae} structure and the \ac{da} regularization, yet still shows small performance improvements over \cite{stephan2020deep}. Fig. \ref{fig:weight_div} visualizes the effects of adding the \ac{da} regularization term. It shows the deviations of the model weights to the model weights of the network trained on the synthetic data over the training epochs. In all cases, the convolutional layers are initialized with the weights from the model trained on the synthetic data. Retraining the network without constraints quickly results in large deviations from the previous model weights. Adding a higher weight regularization factor keeps the model weights closer to the initial weights, therefore ensuring that the model keeps close to the learned general angle estimation without more than a slight drop in performance.

\section{Conclusion}
In this paper, we propose a novel variational autoencoder, whose first layer is drawn from a family of parametric functions that constrain the convolution layers to perform filtering along fast-time and slow-time and separation of different range-Doppler frequencies into distinct kernels similar to conventional FFTs. To overcome the problem of limited radar data, we propose supplementing training through domain adaptation of the network from ray-tracing mathematical model-based synthetic data generated under diverse configurations. We demonstrate the superior performance of the proposed solution compared to conventional radar signal processing and state-of-art neural network solutions for human detection and localization in indoor environments. Apart from superior detection performance, the proposed solution reduces computation requirements drastically by replacing FFTs that consume most operations and simplifying the data flow in the embedded realization by requiring only deep learning accelerators to process raw \ac{adc} data directly.
     
\FloatBarrier
\ifCLASSOPTIONcaptionsoff
  \newpage
\fi

\bibliographystyle{IEEEtran}
\bibliography{SigProcDL}

\end{document}